\documentclass[10pt,twocolumn,letterpaper]{article}

\usepackage[pagenumbers]{cvpr} 

\usepackage[dvipsnames]{xcolor}

\definecolor{cvprblue}{rgb}{0.21,0.49,0.74}
\usepackage[pagebackref,breaklinks,colorlinks,allcolors=cvprblue]{hyperref}
\usepackage{multirow}
\usepackage{soul}
\usepackage{url}
\usepackage{color}
\usepackage{balance} 
\usepackage{xspace}
\usepackage{tabularx}
\usepackage{colortbl}
\usepackage{enumitem}
\usepackage{wasysym}
\usepackage{pifont}
\usepackage{xcolor}
\usepackage{setspace}
\usepackage{algorithm}
\usepackage{amsmath}
\usepackage{algorithm}
\usepackage{algpseudocode}
\usepackage{graphicx}
\usepackage{colortbl}
\usepackage{xcolor}
\usepackage{graphicx}
\usepackage{booktabs}
\usepackage{makecell}
\usepackage[utf8]{inputenc}  
\usepackage[T1]{fontenc}
\usepackage{textcomp}

\definecolor{bblue}{rgb}{0,150,230}
\definecolor{mygray}{gray}{.9}
\definecolor{lightgray}{gray}{.96}
\definecolor{myy}{RGB}{126,95,0}
\definecolor{ggray}{RGB}{127,127,127}
\definecolor{mygreen}{RGB}{93,173,85}
\definecolor{myred}{RGB}{240,16,89}
\definecolor{myblue}{RGB}{0,114,188}
\definecolor{darkgreen}{rgb}{0.0, 0.5, 0.0}
\definecolor{demphcolor}{RGB}{100,100,100}
\definecolor{sh_blue}{rgb}{0,0.60,0.93}
\definecolor{sh_red}{rgb}{0.8627, 0.3098, 0.3176}
\definecolor{lightpink}{rgb}{0.918, 0.761, 0.761}
\definecolor{lightblue}{rgb}{0.671, 0.773, 0.863}
\definecolor{customlightgreen}{rgb}{0.8196, 0.8784, 0.7098}
\definecolor{lightpeach}{rgb}{1.0, 0.882, 0.788}
\definecolor{customblue}{rgb}{0.180, 0.400, 0.522}
\definecolor{lightcyan}{rgb}{0.8196, 0.9725, 0.9804}

\usepackage[normalem]{ulem}

\def\paperID{13059} 
\def\confName{CVPR}
\def\confYear{2025}
\newcommand{\model}{\textit{\textit{TAO}}}
\title{Track Any Anomalous Object: A Granular Video Anomaly Detection Pipeline}
\author{
Yuzhi Huang$^{1}$\textsuperscript{*}, 
Chenxin Li$^{2}$\textsuperscript{*†}, 
Haitao Zhang$^{1}$, 
Zixu Lin$^{1}$, 
Yunlong Lin$^{1}$,\\
Hengyu Liu$^{2}$, 
Wuyang Li$^{2}$, 
Xinyu Liu$^{2}$, 
Jiechao Gao$^{3}$, 
Yue Huang$^{1}$\textsuperscript{†},\\
Xinghao Ding$^{1}$, 
Yixuan Yuan$^{2}$\\
$^1$ Xiamen University \quad
$^2$ The Chinese University of Hong Kong \quad
$^3$ University of Virginia
}

\begin{document}
\maketitle
\footnote{%
\textsuperscript{*} Equal contribution\quad \textsuperscript{†} Corresponding author}

\begin{abstract}
Video anomaly detection (VAD) is crucial in scenarios such as surveillance and autonomous driving, where timely detection of unexpected activities is essential. Albeit existing methods have primarily focused on detecting anomalous objects in videos—either by identifying anomalous frames or objects—they often neglect finer-grained analysis, such as anomalous pixels, which limits their ability to capture a broader range of anomalies. To address this challenge, we propose an innovative VAD framework called \underline{T}rack Any \underline{A}nomalous \underline{O}bject (\model), which introduces a Granular Video Anomaly Detection Framework that, for the first time, integrates the detection of multiple fine-grained anomalous objects into a unified framework. Unlike methods that assign anomaly scores to every pixel at each moment, our approach transforms the problem into pixel-level tracking of anomalous objects. By linking anomaly scores to subsequent tasks such as image segmentation and video tracking, our method eliminates the need for threshold selection and achieves more precise anomaly localization, even in long and challenging video sequences. Experiments on extensive datasets demonstrate that \model~achieves state-of-the-art performance, setting a new progress for VAD by providing a practical, granular, and holistic solution.
For more information, visit the project page at: \href{https://tao-25.github.io/}{https://tao-25.github.io/}
\end{abstract}

\begin{figure}[tb]
  \centering
  \includegraphics[width=\linewidth]{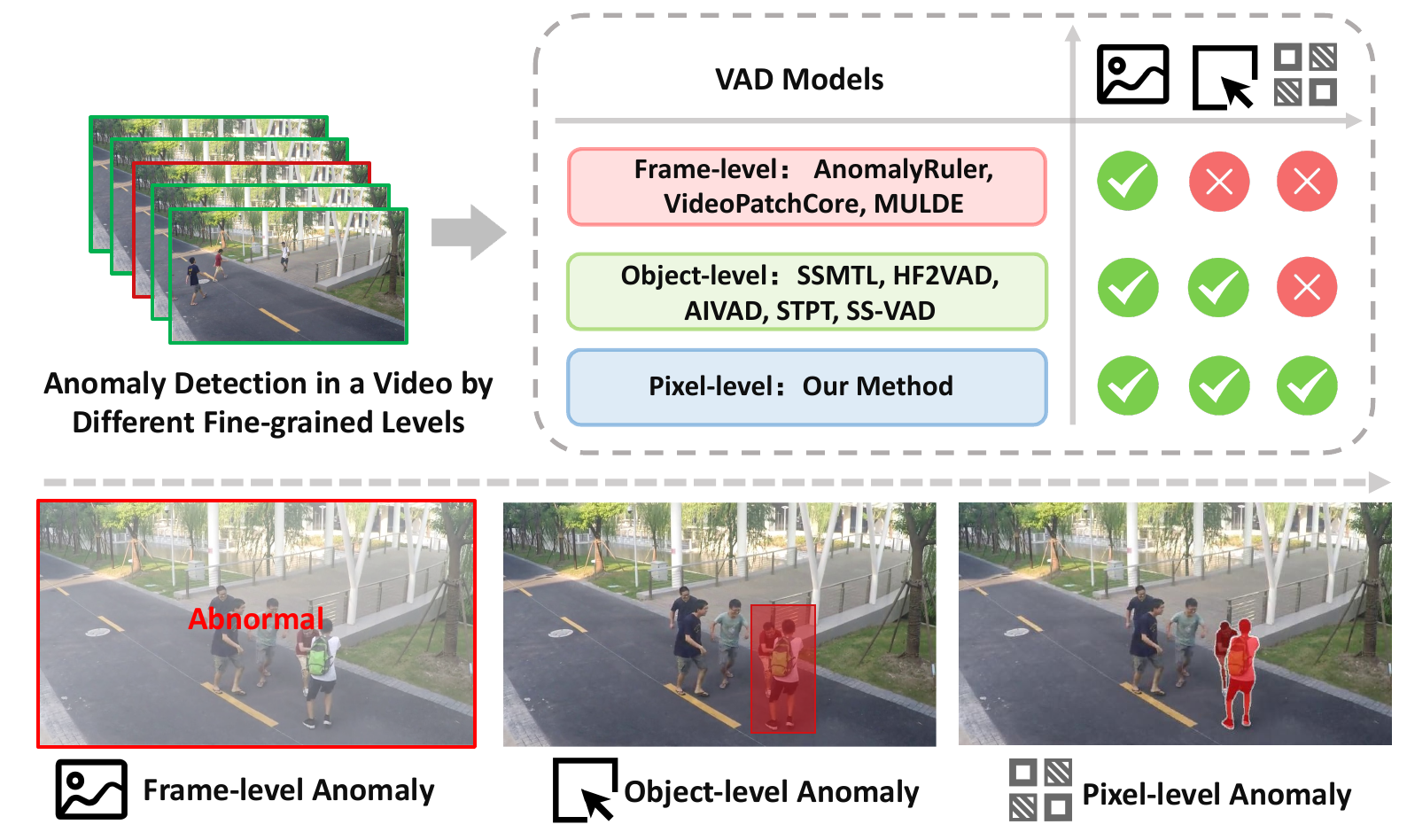}
  \caption{ \textbf{Analyzing the Limitations of Existing VAD Models}. Video anomaly detection (VAD) models are predominantly frame-centric or object-centric. Frame-centric methods detect anomalies in frames without localizing them, while object-centric methods identify anomalous objects but lack pixel-level accuracy. Pixel-centric models address these gaps by providing pixel-level localization, delivering fine-grained segmentation and precise delineation of anomalies, particularly for overlapping objects where traditional methods struggle.
  }
  \label{fig:intro}
\end{figure}

\section{Introduction}
\label{sec:intro}
Video Anomaly Detection (VAD) involves identifying unusual or unexpected activities in surveillance videos and has significant applications in areas such as security monitoring (e.g., detecting violent behavior) and autonomous driving (e.g., recognizing traffic accidents). Current VAD research has developed along two main directions, which are reflected in both methodologies and evaluation benchmarks. Frame-centric methods~\cite{Anomalyruler,videopatchcore,frame1_Hasan_Choi_Neumann_Roy-Chowdhury_Davis_2016,frame2_Park_Noh_Ham_2020,frame3_Shi_Sun_Wu_Jia,frame_5Lo_Oza_Patel_2022,frame4_Liu_Nie_Long_Zhang_Li_2021,bai2023integrating} and their corresponding frame-level benchmarks~\cite{Anomalyruler,mulde,fb_Hinami_Mei_Satoh_2017,wang2025learning} focus on analyzing entire frames to detect global anomalies (e.g., fires or smoke), typically evaluated using frame-level AUC metrics. In contrast, object-centric methods~\cite{AED-SSMTL,AIVAD,OCAD,obj3_Doshi_Yilmaz_2020,BAF-AT,sivl,STPT,bai2024humanedit,wang2023learning} and their object-level benchmarks~\cite{TBDR,AED-SSMTL,OCAD} leverage pre-trained feature extractors to detect object-specific anomalies (e.g., human falls or vehicle accidents), evaluated using metrics such as TBDC~\cite{TBDR} and RBDC~\cite{TBDR}, which comprehensively measure temporal consistency and spatial localization precision.

Despite these advances, a critical gap remains in real-world applications of VAD where fine-grained localization of anomalous objects is essential. As illustrated in Fig.~\ref{fig:intro}, current approaches suffer from fundamental shortage as frame-centric methods can only identify the presence of anomalies without localizing specific regions, while object-centric methods, though more precise, lack pixel-level accuracy. These limitations become particularly pronounced in complex scenarios with overlapping anomalies, where accurate delineation of object boundaries and shapes is crucial~\cite{realtime-weak,tian2021weakly,wu2022self,he2023strategic,he2023degradation,bai2024meissonic,lin2024aglldiff}. This observation motivates us to consider a more challenging yet practical scenario: Can we achieve both object-level structural integrity and pixel-level precision in a granular anomaly detection?

To address this challenge, we propose enhancing existing VAD benchmarks with pixel-level anomaly evaluation. While a straightforward approach would be to employ an anomaly segmentation task for pixel-wise classification~\cite{li2021unsupervised,zhang2021generator,ding2022unsupervised,lin2024fusion2void,he2024diffusion,he2024weakly,fan2024instantsplat}, this method falls short when dealing with multiple, potentially overlapping anomalous objects. Instead, inspired by instance segmentation techniques~\cite{sun2022few,li2022hierarchical,he2025run,xiao2024survey,he2025reti}, we develop a novel evaluation framework that simultaneously considers both object-level and pixel-level anomaly detection accuracy. This framework ensures that detected anomalies accurately correspond to their true shapes, precise positions, and spatial distributions, while effectively minimizing false positives caused by local feature similarities or noise.

However, implementing precise pixel-level tracking in videos presents significant challenges, particularly in maintaining semantic and temporal consistency across frames. Traditional supervised approaches require extensive labeled data, which is scarce in existing datasets with pixel-level annotations. To overcome these challenges, we leverage SAM2~\cite{SAM2}, a large-scale pre-trained segmentation model capable of processing both static images and video streams in real-time without requiring additional fine-tuning on specific anomaly datasets. Its ability to precisely delineate object contours and structures makes it particularly effective in handling complex scenarios involving overlapping anomalies or occlusions.

Building upon these foundations, we propose \model, an integrated framework that combines object-centric anomaly detection algorithms with SAM2 to create a simple yet effective fine-grained video anomaly detection system. For videos containing anomalous frames, our object-centric detection model generates bounding boxes for detected anomalies. After robust filtering, these boxes serve as prompting boxes for SAM2, which then generates segmentation masks for the anomalous objects. Evaluated on our proposed benchmark, our method achieves state-of-the-art performance on two widely-used datasets, demonstrating its effectiveness in bridging the gap between object-level detection and pixel-precise segmentation. The contributions of our paper are as follows:
\begin{itemize}
\item[$\bullet$] We present a new testing standard that unifies pixel-level and object-level assessment in video anomaly detection, addressing the limitations of existing metrics in complex scenarios with multiple or overlapping anomalies.
\item[$\bullet$] We introduce \model, a streamlined framework that integrates object-centric anomaly detection with the segmentation capabilities from vision foundation model, enabling precise pixel-level tracking of anomalous objects without requiring additional fine-tuning.
\item[$\bullet$] Through extensive experiments on UCSD Ped2 and ShanghaiTech Campus datasets, we demonstrate that our approach achieves state-of-the-art performance under both traditional metrics and our proposed benchmark.
\end{itemize}

\section{Related Work}
\label{sec:related_work}
\begin{figure*}[t!]
    \centering
    \includegraphics[width=\linewidth]{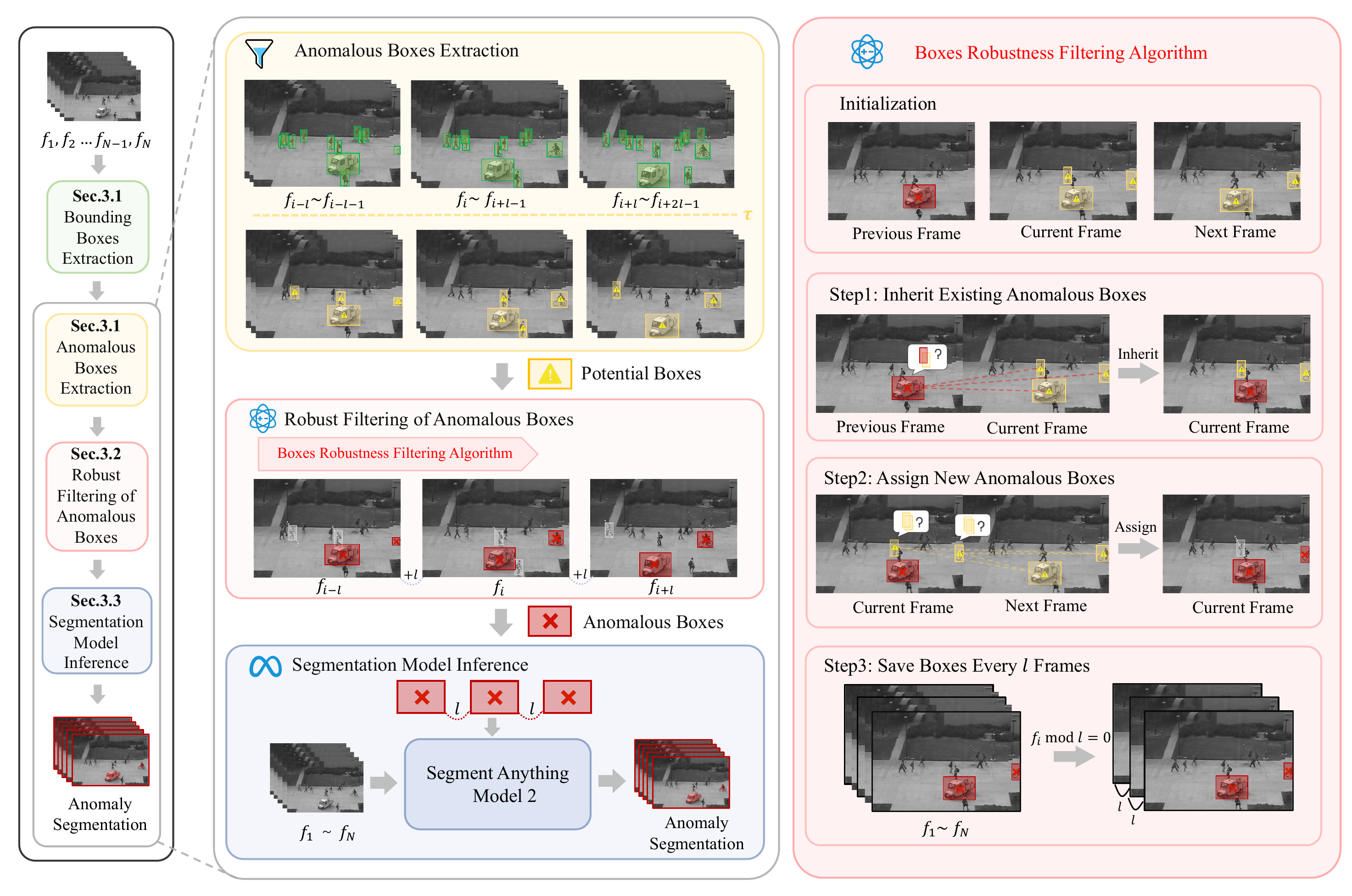}
    \caption{\textbf{Pipeline of our proposed \model.} We first generate bounding boxes to identify objects in each frame. Next, we score these boxes using object-centric video anomaly detection algorithms to extract potential anomalous boxes. To ensure robustness, we apply filtering to eliminate redundant boxes. Finally, the filtered boxes and original frames are input into a prompt-based segmentation model to produce pixel-level anomaly segmentation masks. }
    \label{fig:method1}
\end{figure*}

\noindent \textbf{Video Anomaly Detection.}  Video anomaly detection (VAD) presents significant challenges due to the rarity of anomaly data and the wide variety of abnormal events, which hinder the generalization capabilities of existing models across diverse scenarios. Traditional VAD methods can be broadly classified into frame-centric and object-centric approaches.
Frame-centric methods~\cite{Anomalyruler,videopatchcore,frame1_Hasan_Choi_Neumann_Roy-Chowdhury_Davis_2016,frame2_Park_Noh_Ham_2020,frame3_Shi_Sun_Wu_Jia,frame_5Lo_Oza_Patel_2022,frame4_Liu_Nie_Long_Zhang_Li_2021,xing2024segmamba,xing2024cross} analyze entire frames or sequences, often leveraging reconstruction or prediction errors to identify anomalies. These methods are particularly effective for detecting global events, such as fires or smoke. However, they struggle with localized anomalies and are prone to interference from normal regions within a frame, limiting their precision in complex scenarios ~\cite{frame_5Lo_Oza_Patel_2022,frame1_Hasan_Choi_Neumann_Roy-Chowdhury_Davis_2016,frame2_Park_Noh_Ham_2020,frame4_Liu_Nie_Long_Zhang_Li_2021,xing2022nestedformer}.
Object-centric methods~\cite{AED-SSMTL,AIVAD,OCAD,obj2_Doshi_Yilmaz_2020,obj3_Doshi_Yilmaz_2020,BAF-AT,sivl,obj4_Yu_Wang_Cai_Zhu_Xu_Yin_Kloft_2020,xing2023diff} focus on specific objects within frames by employing pre-trained object detection models to extract bounding boxes and assess their abnormality. This targeted approach is better suited for detecting anomalies related to individuals or objects, such as human falls or vehicle accidents. By concentrating on specific regions and reducing redundant information, object-centric methods achieve improved accuracy and robustness, especially in complex, multi-object environments.
Building on these foundations, our approach is the first to integrate large-scale pre-trained models for pixel-level fine-grained detection in VAD. This advancement enables precise anomaly localization, bridging the gap between frame-level analysis and object-level detection, and significantly enhancing overall detection performance.

\noindent \textbf{Vision Foundation Models.} In recent years, large-scale pre-trained language models (LLMs) and vision-language models (VLMs) have shown remarkable potential in advancing video anomaly detection. Models such as BLIP-2~\cite{blip-2}, LLaVA~\cite{llava}, LAVAD~\cite{zanella2024harnessing}, VadCLIP~\cite{vadclip}, Video-ChatGPT~\cite{maaz2023video}, and Video-LLaMA~\cite{video-llama} have significantly improved the understanding of complex visual tasks by integrating visual and linguistic information. Concurrently, prompting techniques have gained traction in large-scale vision models, enabling substantial progress in tasks such as image segmentation. By employing semantic prompts (e.g., free-form text) and spatial prompts (e.g., points or bounding boxes), these models achieve highly accurate segmentation guided by input cues.
Segmentation Foundation Models (SFMs), such as the Segment Anything Model (SAM)~\cite{SAM} and SEEM~\cite{SEEM}, have demonstrated exceptional performance, particularly in zero-shot generalization and multi-modal prompting, allowing for cross-task segmentation. Building on these advancements, the Segment Anything Model 2 (SAM2)~\cite{SAM2} further improves segmentation, particularly for video-based applications. SAM2 integrates prompting techniques to not only segment static objects with precision but also capture dynamic, complex objects in video contexts. This capability enables SAM2 to deliver finer segmentation and enhanced localization accuracy in video anomaly detection, addressing the challenges posed by real-world variability and complexity.
This work leverages SAM2's advanced video segmentation capabilities to tackle the challenge of fine-grained video anomaly detection, achieving precise and robust anomaly identification in complex scenarios.

\section{Method}
\label{sec:method}

\begin{figure*}[t]
  \centering
  \includegraphics[width=\linewidth]{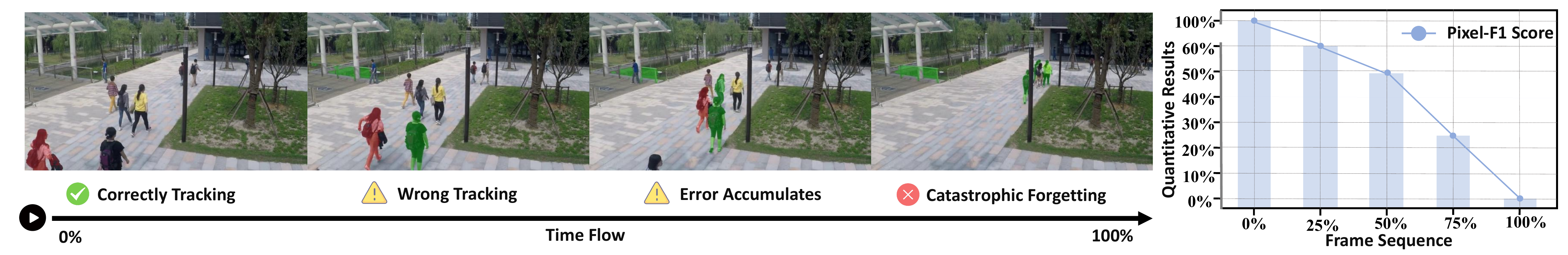}
  \caption{ \textbf{Impact of Redundant Segmentation on Performance}. 
  Potential anomalous boxes are used as prompts for SAM2 segmentation (red masks: true anomalies, green masks: redundant results). Initially, SAM2 tracks true anomalies accurately, but tracking errors accumulate, leading to catastrophic forgetting and a collapse in Pixel-F1 performance. 
  }
  \label{fig:method2}
\end{figure*}

\begin{algorithm}[t]
\caption{Boxes Robustness Filtering}
\label{alg:robust_bounding_box}
\begin{algorithmic}[1]
    \State \textbf{Input:} A set of anomalous bounding boxes $\mathcal{B}_{\text{anomaly}} = \{b_i \mid s_i > \tau\}$ for each frame $f_i \in \{f_1, f_2, \dots, f_n\}$, where $s_i$ is the anomaly score and $\tau$ is the threshold; parameters $k$ (tracking window size), $h$ (overlap threshold), $l$ (save interval), and $m$ (frame match threshold)
    \vspace{-1pt} 
    \State \textbf{Output:} Filtered bounding boxes $\mathcal{B}_{\text{filtered}}$ with their corresponding object labels $\mathcal{L}_j$
    \vspace{-1pt}
    \State Initialize the list $\mathcal{B}_{i}$ to store the filtered bounding boxes from frame $f_i$
    \vspace{-1pt}
    \For{each frame $f_i \in \{f_1, \dots, f_n\}$}
        \vspace{-1pt}
        \State \textbf{Step 1: Inherit Existing Anomalous Boxes}
        \For{each bounding box $b_j \in \mathcal{B}_{\text{anomaly}}$ and $b_p$ from frame $f_{i-k}$ to $f_{i-1}$}
            \If{$\sum \limits_{p=i-k}^{i-1} \mathbb{I}(\text{IoU}(b_j, b_p) > h) \geq m$}
                \State Assign $\mathcal{L}_j = \mathcal{L}_p$
                \State Remove $b_j$ from $\mathcal{B}_{i}$
            \EndIf
        \EndFor
        \vspace{-1pt}
        \State \textbf{Step 2: Assign New Anomalous Boxes}
        \For{each remaining $b_j \in \mathcal{B}_{\text{anomaly}}$ and $b_p$ from frame $f_{i+1}$ to $f_{i+k}$}
            \If{$\sum \limits_{p=i+1}^{i+k} \mathbb{I}(\text{IoU}(b_j, b_p) > h) \geq m$}
                \State Assign a new object label $\mathcal{L}_j = \mathcal{F}(b_j)$
                \State Save the tuple $\mathcal{T}_{\text{box}} = (f_i, b_j, \mathcal{L}_j)$
            \EndIf
        \EndFor
        \vspace{-1pt}
        \State \textbf{Step 3: Save Boxes Every $l$ Frames}
        \If{$i \bmod l == 0$}
            \State Save all bounding box tuples $\mathcal{T}_{\text{box}} = (f_i, b_j, \mathcal{L}_j)$ in the current frame
        \EndIf
        \vspace{-1pt}
    \EndFor
\end{algorithmic}
\end{algorithm}

\noindent\textbf{Preliminary.}
In the proposed video anomaly detection (VAD) benchmark, a video clip is represented as a sequence \(\{f_1, f_2, \dots, f_{N_f}\} \subseteq \mathcal{C}\), where \(\mathcal{C}\) is the set of all possible video clips, and \(N_f\) denotes the total number of frames. Each frame \(f_i\) is expressed as \(f_i = [p_{i,1}, p_{i,2}, \dots, p_{i,N_i}]\), where \(N_i\) is the total number of pixels in the frame, and \(p_{i,j} \in \mathcal{P}\) represents the \(j\)-th pixel, with \(\mathcal{P}\) as the set of possible pixel values. Each pixel \(p_{i,j}\) is classified as either ``normal'' or ``anomalous''.

\noindent\textbf{Overview.}
To achieve precise anomaly localization in videos, we design a streamlined pipeline for pixel-level segmentation in video anomaly detection. Our pipeline comprises four stages: bounding boxes extraction, anomalous boxes extraction, robust filtering, and segmentation inference. The process begins by generating bounding boxes to identify objects of interest. These boxes are then scored using object-centric anomaly detection algorithms to extract potential anomalies. A tracking-based filtering step follows to eliminate redundant boxes and retain only robust ones. Finally, the filtered boxes and original frames are fed into a prompt-based segmentation model to generate pixel-level anomaly masks. The pipeline is illustrated in Fig.~\ref{fig:method1}.


\subsection{Anomalous Boxes Extraction}
\label{sec:4.3}
\noindent\textbf{Bounding Boxes Extraction.}
To achieve precise pixel-level segmentation in video anomaly detection, pre-trained object detection algorithms are used to generate bounding boxes for objects within a frame, serving as potential prompts for guiding the segmentation model. For a given input frame $f$, the object detection model $\mathcal{D}$ outputs $m$ bounding boxes $b_1, b_2, \dots, b_m$, each paired with a class label $y_1, y_2, \dots, y_m$:

\vspace{-5pt}

\begin{equation}
\{(b_1, y_1), (b_2, y_2), \dots, (b_m, y_m)\} = \mathcal{D}(f)
\end{equation}

After the extraction of object bounding boxes in each frame, object-centric VAD algorithms compute an anomaly score \(s_i\) for each object within its bounding box \(b_i\). This score is derived from features such as pose, speed, and depth, and can be formally represented as:
\begin{equation}
s_i = \mathcal{A}(b_i)
\end{equation}
where \(\mathcal{A}\) represents the scoring function used in object-centric video anomaly detection algorithms.
To isolate anomalous objects, we filter the bounding boxes based on an anomaly score threshold. Specifically, we retain only the bounding boxes where the anomaly score exceeds a predefined threshold \( \tau \), as shown in the following expression:
\begin{equation}
\mathcal{B}_{\text{anomaly}} = \{b_i \mid s_i > \tau \}
\end{equation}
where \( \tau \) denotes the anomaly score threshold, \( s_i \) is the anomaly score for the \( i \)-th object, and \( b_i \) is the corresponding bounding box.

\subsection{Robust Filtering of Anomalous Boxes}
\label{sec:4.4}
The anomalous boxes detection from Sec.~\ref{sec:4.3} contains numerous false positives due to overlapping characteristics between normal and abnormal boxes. This overlap complicates threshold setting and leads to misclassifications. These redundant boxes can significantly degrade segmentation performance by generating inaccurate results and overwhelming the model's tracking capabilities, potentially causing the loss of anomalous targets in subsequent frames.

As illustrated in Fig.~\ref{fig:method2}, using potential anomalous boxes as SAM2 prompts at fixed intervals reveals the impact of tracking errors. While red masks indicate true anomalies and green masks show redundant results, the model initially tracks anomalous objects effectively. However, tracking errors accumulate over time, leading to an increasing number of incorrectly tracked objects. This deterioration ultimately results in catastrophic forgetting of true anomalies, reflected in steadily declining Pixel-F1 scores.

To enhance robustness and address redundancy, we present the \textit{Boxes Robustness Filtering} algorithm, which exploits the temporal consistency of true anomalous boxes. Unlike redundant boxes that appear sporadically, true anomalous boxes maintain consistent tracking of the same target. The algorithm, detailed in Alg.~\ref{alg:robust_bounding_box}, operates in three key steps.
First, the \textit{inheritance step} computes the similarity between each bounding box \(b_j\) in the current frame \(f_i\) and bounding boxes \(b_p\) from the previous \(k\) frames. If the Intersection over Union (IoU) exceeds a threshold \(h\) for at least \(m\) frames, \(b_j\) inherits the label \(\mathcal{L}_p\) from earlier frames, ensuring consistent tracking of previously identified anomalies. Second, the \textit{assignment step} addresses bounding boxes that fail to inherit a label by comparing them with bounding boxes in the subsequent \(k\) frames. If the IoU exceeds \(h\) for at least \(m\) frames, a new label \(\mathcal{L}_j = \mathcal{F}(b_j)\) is assigned, effectively labeling newly emerging anomalies. Finally, the \textit{saving step} records bounding boxes and their labels every \(l\)-th frame, systematically capturing all identified anomalies.
By iteratively applying these steps, the \textit{Boxes Robustness Filtering} algorithm achieves robust tracking and localization of anomalous objects. Each bounding box and its label are stored as a tuple \(\mathcal{T}_{\text{box}} = (f_i, b_j, \mathcal{L}_j)\), ensuring spatial and temporal consistency across video frames. This approach reduces redundancy and strengthens anomaly detection accuracy, providing a reliable basis for subsequent segmentation and analysis.

\subsection{Inference of the Segmentation Model}
Using the tuple \(\mathcal{T}_{\text{box}} = (f_i, b_j, \mathcal{L}_j)\) obtained from Sec.~\ref{sec:4.4}, we extract the center point of each bounding box \(b_j\). Specifically, given the coordinates of the box \(b_j = [x_1, y_1, x_2, y_2]\), the center point is calculated:
$\mathbf{c}_j = \left( \frac{x_1 + x_2}{2}, \frac{y_1 + y_2}{2} \right)$.
Effectively, these prompts are not saved for every frame but are instead collected at regular intervals, specifically every \(l\)-th frame, as outlined in Sec.~\ref{sec:4.4}. This ensures efficient storage and processing while still capturing the necessary information for accurate segmentation~\cite{li2024u,li2021consistent}. For each selected frame, both the center point \(\mathbf{c}_j\) and the bounding box \(b_j\) are stored as prompts. Rather than processing each frame individually, we aggregate prompts from all saved frames \(\{f_1, f_{1+l}, f_{1+2l}, \dots, f_{N_f}\}\), where \(l\) denotes the saving interval. The aggregated prompts are:
\begin{equation}
\mathcal{I} = \{(\mathbf{c}_j, b_j, f_i) \mid i = 1, 1+l, 1+2l, \dots, N_f\}
\end{equation}
where \(\mathcal{I}\) represents the set of prompts that includes the center points, bounding box coordinates \(b_j\), and original frames \(f_i\) for the selected frames.

By inputting this prompt information \(\mathcal{I}\) saved at regular intervals into SAM2, which serves as the prompt-based segmentation model \(\mathcal{M}\), we leverage its feature propagation capabilities to efficiently generate pixel-level segmentation results for the entire video. SAM2 propagates the segmentation results from the prompted frames to the remaining frames, producing pixel-level segmentation across the entire video:
\begin{equation}
  \mathcal{S} = \mathcal{M}(\mathcal{I}) = \{\mathcal{S}_1, \mathcal{S}_2, \dots, \mathcal{S}_{N_f}\}  
\end{equation}
where \(\mathcal{S}_i = [p_1, p_2, \dots, p_M]\) denotes the segmentation result for frame \(f_i\), where each pixel \(p_j\) is either ``normal'' or ``anomalous''. This approach enables the generation of the complete segmentation set \(\mathcal{S}\), efficiently utilizing contextual information from the prompts.

\section{Experiments}
\label{sec:experiments}

\begin{figure*}[t]
    \centering
    \caption{Qualitative comparison with anomaly detection models adapted for video anomaly detection on the UCSD Ped2 dataset.}
    \includegraphics[width=\linewidth]{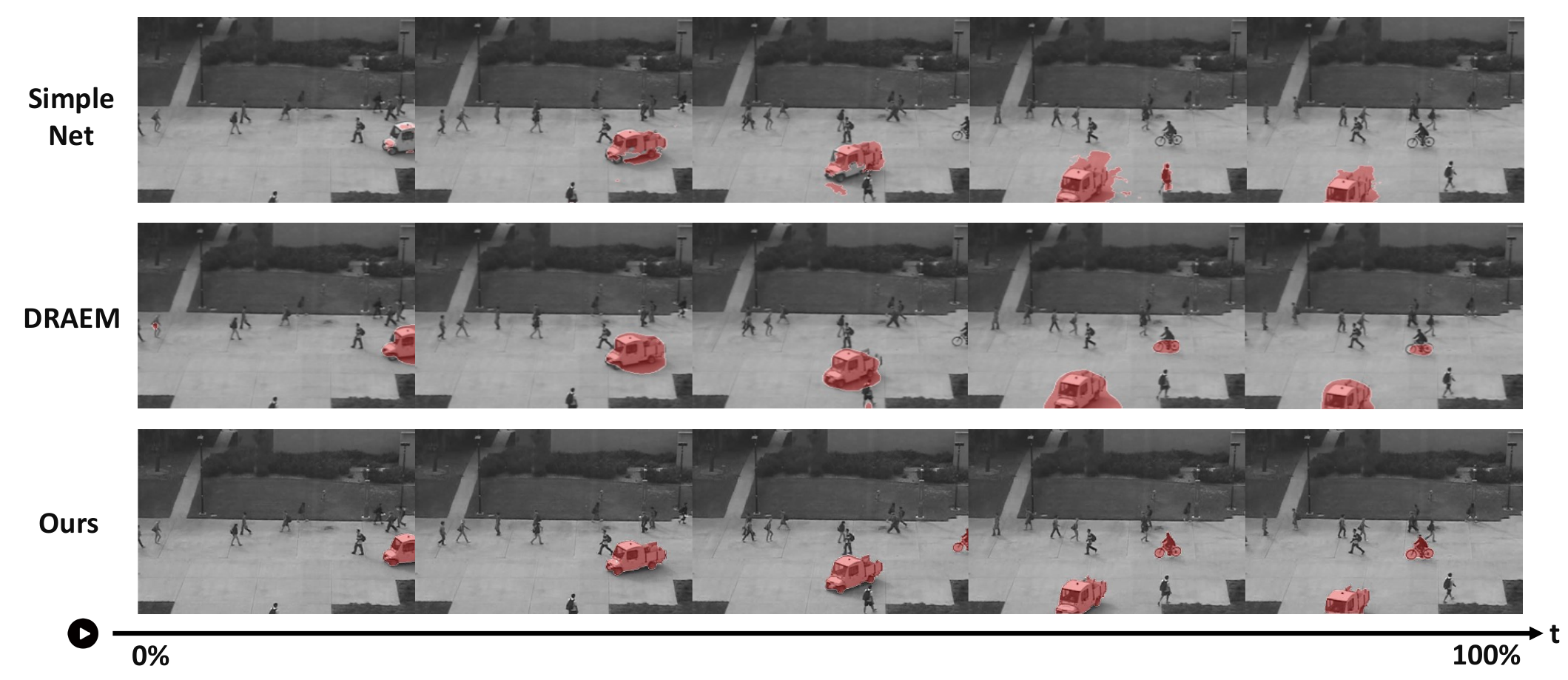}

    \label{fig:experiments}
\end{figure*}

\begin{table*}[t]
\centering
\caption{Quantitative comparison on video anomaly detection with pixel-level and object-level metrics on the UCSD Ped2 dataset.}
\label{tab1}
\resizebox{0.95\textwidth}{!}{
\begin{tabular}{l!{\vrule width 1.2pt}cccc!{\vrule width 1.2pt}cc}
\toprule
& \multicolumn{4}{c!{\vrule width 1.2pt}}{\textbf{Pixel-level}} & \multicolumn{2}{c}{\textbf{Object-level}} \\
\cmidrule(lr{1.2pt}){2-5} \cmidrule(lr{1.2pt}){6-7} 
\textbf{Method} & \textbf{Pixel-AUROC ↑} & \textbf{Pixel-AP ↑} & \textbf{Pixel-AUPRO ↑} & \textbf{Pixel-F1 ↑} & \textbf{RBDC ↑} & \textbf{TBDC ↑} \\
\midrule
AdaCLIP (Zero-shot)~\cite{AdaCLIP} & 51.02 & 1.32 & 33.98 & 2.61 & 5.8 & 10.6 \\
AdaCLIP (Fully fine-tuned)~\cite{AdaCLIP} & 53.06 & 4.97 & \cellcolor{orange!30}50.66 & \cellcolor{orange!30}11.19 & 12.3 & 15.5 \\
AnomalyCLIP (Zero-shot)~\cite{Anomalyclip} & 51.63 & 21.20 & 36.34 & 5.92 & 7.5 & 11.2 \\
AnomalyCLIP (Fully fine-tuned)~\cite{Anomalyclip} & 54.25 & \cellcolor{yellow!30}23.73 & 38.59 & 7.48 & 13.1 & 21.0 \\
DDAD (Fully trained)~\cite{DDAD} & \cellcolor{yellow!30}55.87 & 5.61 & 15.12 & 2.67 & 18.01 & 13.29 \\
SimpleNet (Fully trained)~\cite{SimpleNet} & 52.49 & 20.51 & \cellcolor{yellow!30}44.05 & 10.71 & \cellcolor{orange!30}51.18 & \cellcolor{yellow!30}27.75 \\
DRAEM (Fully trained)~\cite{DRAEM} & \cellcolor{orange!30}69.58 & \cellcolor{orange!30}30.63 & 35.78 & \cellcolor{yellow!30}10.89 & \cellcolor{yellow!30}44.26 & \cellcolor{orange!30}70.64 \\
\rowcolor{red!5} \model~(Partially fine-tuned) & \cellcolor{red!30}75.11 & \cellcolor{red!30}50.78 & \cellcolor{red!30}72.97 & \cellcolor{red!30}64.12 & \cellcolor{red!30}83.6 & \cellcolor{red!30}93.2 \\
\bottomrule
\end{tabular}
}
\end{table*}

\begin{table}[tb]
\centering
\caption{Quantitative comparison on video anomaly detection with RBDC and TBDC metrics on the UCSD Ped2 dataset.}
\label{tab2}
\begin{tabular}{lcc}
\toprule
\textbf{Method} & \textbf{RBDC ↑} & \textbf{TBDC ↑} \\
\midrule

OCAD~\cite{OCAD} & 52.7 & 72.8 \\
SS-VAD~\cite{TBDR} & 62.5 & 80.5 \\
SiVL~\cite{sivl} & \cellcolor{orange!30}74 & \cellcolor{yellow!30}89.3 \\
AED-SSMTL~\cite{AED-SSMTL} & \cellcolor{yellow!30}72.8 & \cellcolor{orange!30}91.2 \\
\rowcolor{red!5}\model~(Ours) & \cellcolor{red!30}83.6 & \cellcolor{red!30}93.2 \\
\bottomrule
\end{tabular}
\end{table}

\begin{table}[t]
\centering
\caption{Quantitative comparison on video anomaly detection with RBDC and TBDC metrics on the ShanghaiTech Campus dataset.}
\label{tab3}
\begin{tabular}{lcc}
\toprule
\textbf{Method} & \textbf{RBDC ↑} & \textbf{TBDC ↑} \\
\midrule
OCAD~\cite{OCAD} & 20.7 & 44.5 \\
BAF-AT~\cite{BAF-AT} & 41.3 & 78.8 \\
AED-SSMTL~\cite{AED-SSMTL} & 43.2 & 84.1 \\
HF2VAD~\cite{HF2VAD} & \cellcolor{yellow!30}45.4 & \cellcolor{yellow!30}84.5 \\
STPT~\cite{STPT} & \cellcolor{orange!30}51.6 & \cellcolor{orange!30}84.6 \\
\rowcolor{red!5}\model~(Ours) & \cellcolor{red!30}62.1 & \cellcolor{red!30}85.4 \\
\bottomrule
\end{tabular}
\end{table}

\subsection{A New Comprehensive VAD Benchmark}
\noindent\textbf{Benchmark Design.}
We present a novel benchmark framework that integrates both pixel-level and object-level evaluation metrics to advance VAD model assessment. This dual-level approach addresses the limitations of using single evaluation metrics. \textit{Pixel-level metrics} excel at identifying subtle irregularities through independent pixel evaluation, but struggle to capture broader spatial patterns. \textit{Object-level metrics} complement this by assessing global characteristics like shape and position, enabling coherent anomaly tracking and recognition. By combining these complementary perspectives, our framework provides a comprehensive evaluation of both fine-grained details and holistic anomaly patterns, particularly valuable for real-world applications.

\noindent\textbf{Benchmark Datasets.}
\textit{UCSD Ped2}~\cite{ped2} contains 16 training and 12 test videos (240×360 pixels) captured by a fixed overhead camera. The training set consists of normal pedestrian activities, while the test set includes anomalous events such as bikers, skateboarders, and vehicles. \textit{ShanghaiTech Campus}~\cite{shanghaitech}, one of the largest VAD datasets, comprises 330 training and 107 test videos (480×856 pixels). While the training set features normal scenarios, the test set contains various anomalies including robbery, fighting, and unauthorized cycling in pedestrian zones.

\noindent\textbf{Benchmark Metrics.}
For \textit{pixel-level} evaluation, we employ four metrics: Pixel-AUROC measures discrimination between normal and anomalous pixels across thresholds, Pixel-AP assesses detection precision by balancing false positives and negatives, Pixel-AUPRO quantifies segmentation accuracy through region overlap, and Pixel-F1 provides an integrated precision-recall measure. For \textit{object-level} assessment, we utilize Region-Based Detection Criterion (RBDC) and Track-Based Detection Criterion (TBDC). RBDC measures spatial accuracy using Intersection over Union (IoU) with threshold $\alpha$, while TBDC evaluates temporal consistency in tracking anomalous regions across frames. Together, these metrics provide comprehensive evaluation of both spatial precision and temporal coherence.


\subsection{Implementation Details}
Our framework builds upon SAM2~\cite{SAM2}, utilizing an object-level anomaly detection algorithm~\cite{AIVAD} to extract anomalous bounding boxes as segmentation prompts. We employ SAM2~\cite{SAM2} as the prompt-based segmentation model, which integrates these prompts with original frames for precise anomaly segmentation. Our implementation adopts a {partial fine-tuning} strategy, where only the object-level VAD algorithm is fine-tuned on target datasets, while SAM2 retains its pre-trained weights for inference. In the bounding box threshold filtering stage, we assign anomaly scores based on pose and depth features, with thresholds set to $\tau = 1.5$ for UCSD Ped2 and $\tau = 1.6$ for ShanghaiTech Campus. The subsequent robustness filtering stage employs a tracking window of $k = 5$, frame match threshold of $m = 3$, and box overlap threshold of $h = 0.2$, with save intervals of $l = 5$ and $l = 15$ for UCSD Ped2 and ShanghaiTech Campus respectively. For segmentation, we utilize SAM2 with hiera base-plus weights on an NVIDIA RTX 4090 GPU. To ensure robust evaluation, overlapping boxes of the same anomaly are merged, and segmentation probability maps are uniformly binarized. For object-level metrics, we derive bounding boxes from segmentation extremal coordinates. Given anomaly pixels $P = {(x_1, y_1), (x_2, y_2), \dots, (x_N, y_N)}$, we compute the box coordinates as:
$x_{\text{min}} = \min_{(x_i, y_i) \in P} x_i, \quad x_{\text{max}} = \max_{(x_i, y_i) \in P} x_i,
y_{\text{min}} = \min_{(x_i, y_i) \in P} y_i, \quad y_{\text{max}} = \max_{(x_i, y_i) \in P} y_i$,
where the resulting bounding box $(x_{\text{min}}, y_{\text{min}}, x_{\text{max}}, y_{\text{max}})$ is used for computing RBDC and TBDC metrics, providing robust spatial evaluation at the object level.

\begin{table*}[tb]
\centering
\caption{Ablation study on proposed key components, evaluated with pixel-level and object-level metrics on the UCSD Ped2 dataset.}
\label{tab4}
\resizebox{0.95\textwidth}{!}{%
\begin{tabular}{cc|c|c|c|c|c|c}
\toprule
& & \multicolumn{4}{c|}{\textbf{Pixel-level}} & \multicolumn{2}{c}{\textbf{Object-level}} \\
\cmidrule(lr){3-6} \cmidrule(lr){7-8}
\textbf{\makecell{Video Track\\ Mode}} & \textbf{\makecell{Box Robust \\Filtering Algorithm}} & \textbf{Pixel-AUROC ↑} & \textbf{Pixel-AP ↑} & \textbf{Pixel-AUPRO ↑} & \textbf{Pixel-F1 ↑} & \textbf{RBDC ↑} & \textbf{TBDC ↑} \\
\midrule
× & × & \cellcolor{yellow!30}41.72 & \cellcolor{yellow!30}25.30 & \cellcolor{yellow!30}40.55 & \cellcolor{yellow!30}32.99 & \cellcolor{yellow!30}41.0 & 71.6 \\
× & \checkmark & 39.06 & 22.80 & 38.09 & 30.56 & 39.3 & \cellcolor{yellow!30}71.7 \\
\checkmark & × & \cellcolor{orange!30}68.90 & \cellcolor{orange!30}38.47 & \cellcolor{orange!30}67.23 & \cellcolor{orange!30}52.17 & \cellcolor{orange!30}70.8 & \cellcolor{orange!30}76.9 \\
\checkmark & \checkmark & \cellcolor{red!30}75.11 & \cellcolor{red!30}50.78 & \cellcolor{red!30}72.97 & \cellcolor{red!30}64.12 & \cellcolor{red!30}83.6 & \cellcolor{red!30}93.2 \\
\bottomrule
\end{tabular}%
}
\end{table*}

\subsection{Comparison to Anomaly Detection Models}
Given the lack of pixel-level segmentation capabilities in current video anomaly detection models, we benchmark several state-of-the-art image anomaly detection algorithms, including AdaCLIP~\cite{AdaCLIP}, AnomalyCLIP~\cite{Anomalyclip}, DRAEM~\cite{DRAEM}, DDAD~\cite{DDAD}, and SimpleNet~\cite{SimpleNet}, to evaluate both pixel-level and object-level metrics on the UCSD Ped2 dataset. AdaCLIP and AnomalyCLIP were selected for their pre-trained vision-language architectures, which closely align with SAM2’s design and demonstrate strong zero-shot capabilities. To ensure a fair comparison, we assess both AdaCLIP and AnomalyCLIP under zero-shot and fully fine-tuned settings. In contrast, DRAEM, DDAD, and SimpleNet, as fully trained models, provide robust baselines for pixel-level and object-level anomaly detection, representing diverse methodological approaches.
For anomaly detection, video streams are processed frame-by-frame through each algorithm, generating heatmaps that identify potential anomalies. Pixel-level segmentation is achieved by applying a threshold to these heatmaps, allowing precise localization of anomalous regions.

As shown in Tab.\ref{tab1}, the experimental results clearly demonstrate the advantages of our model in both pixel-level and object-level anomaly detection. While most benchmarked models struggle to balance these two aspects, often excelling in one while underperforming in the other, our model achieves robust performance across both dimensions. This balance reflects its strong and versatile design, enabling precise pixel-level segmentation and maintaining spatial-temporal consistency at the object level. Moreover, our partial fine-tuning approach effectively leverages pre-trained features, minimizing the need for extensive dataset-specific training. In contrast, fully trained models such as DRAEM, DDAD, and SimpleNet, as well as vision-language models like AdaCLIP and AnomalyCLIP, fail to achieve comparable results. Fig.~\ref{fig:experiments} further illustrates the strengths of our method. The segmentation results produced by our model are more complete and detailed, accurately delineating object boundaries while preserving the integrity of anomalous objects.

\subsection{Comparsion to Conventional Video Anomaly Detection Models}
To assess the performance of our proposed method, we conducted comprehensive experiments on two widely used datasets, UCSD Ped2 and ShanghaiTech Campus. Our approach was benchmarked against several state-of-the-art video anomaly detection models, including SS-VAD~\cite{TBDR}, SiVL~\cite{sivl}, AED-SSMTL~\cite{AED-SSMTL}, HF2VAD~\cite{HF2VAD}, BAF-AT \cite{BAF-AT}, and STPT \cite{STPT}. Due to the absence of fine-grained segmentation capabilities in traditional methods, the evaluation was focused on object-centric anomaly detection and limited to object-level metrics, TBDC~\cite{TBDR} and RBDC~\cite{TBDR}. These metrics provide a robust and meaningful basis for comparing the effectiveness of our model with existing approaches.

Our method achieves state-of-the-art performance on both the UCSD Ped2 and ShanghaiTech Campus datasets. On UCSD Ped2, it outperforms existing methods in TBDC and RBDC metrics, demonstrating precise tracking and localization of anomalous objects across video frames. On the more challenging ShanghaiTech Campus dataset, characterized by complex backgrounds and subtle anomalies, our method consistently surpasses baselines by delivering balanced performance across all evaluation metrics. Unlike existing approaches, which often excel in isolated metrics but fail to generalize, our model effectively integrates object-centric detection and fine-grained segmentation, enabling robust anomaly localization and tracking under diverse real-world conditions, as shown in Tabs.~\ref{tab2} and~\ref{tab3}.

To conclude, our method establishes a new standard in video anomaly detection by combining object-centric detection with fine-grained segmentation. This integration enables precise localization and robust tracking of anomalous objects, addressing the limitations of traditional methods that struggle with balanced performance across metrics. Our approach demonstrates versatility and effectiveness across diverse datasets, highlighting its potential for real-world applications.

\begin{figure}[h]
\centering
\includegraphics[width=\linewidth]{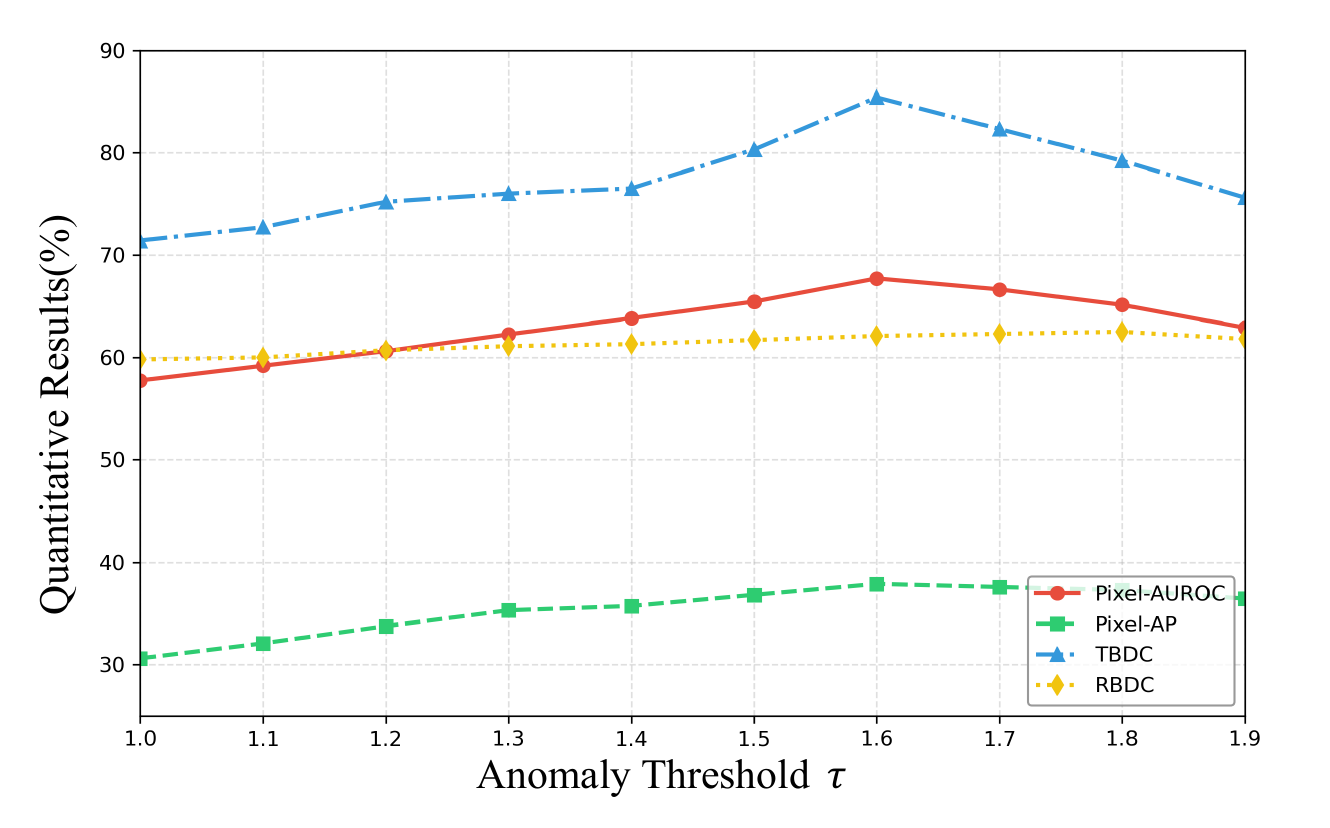}
\caption{\textbf{Sensitivity  Analysis of Anomaly Threshold ($\tau$).} We evaluate performance under varying threshold on ShanghaiTech Campus dataset. 
}
\label{fig:experiments2}
\end{figure}

\begin{figure}[h]
\centering
\includegraphics[width=\linewidth]{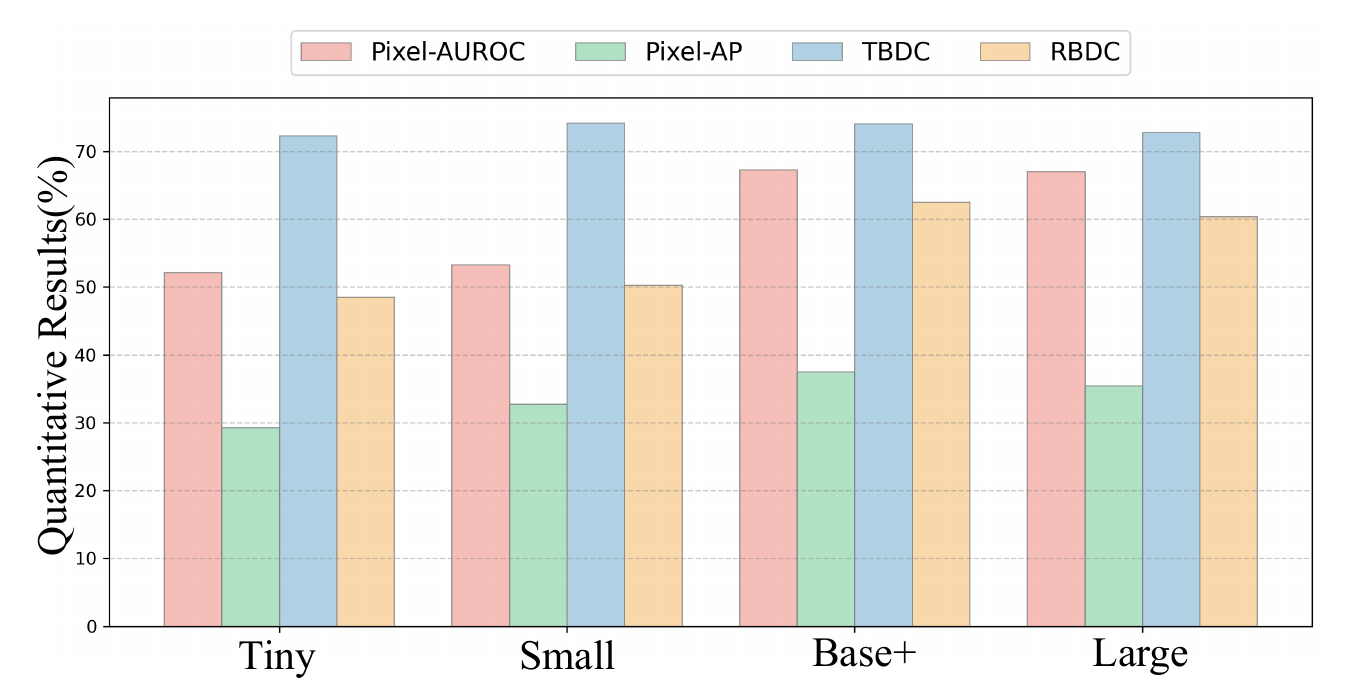}
\caption{\textbf{Robust Testing of SAM2 Backbones.} We compare the performance across different SAM2 model variants. 
}
\label{fig:experiments3}
\end{figure}

\subsection{
Further Empirical Study}

\noindent\textbf{Ablation Study.}
We conduct ablation experiments on the UCSD Ped2 dataset to evaluate two components: \textit{video tracking mode} and \textit{box robust filtering}. Video tracking mode, a SAM2 component, enables temporal analysis, while disabling it reduces the model to SAM, performing frame-by-frame segmentation. Disabling box robust filtering uses raw anomaly-thresholded boxes as prompts.
Tab.~\ref{tab4} shows the effects of each component on pixel- and object-level metrics. The baseline model, with both components disabled, performs the worst. Enabling box robust filtering alone yields minimal improvement, while activating video tracking mode significantly boosts pixel-level accuracy. The best performance is achieved with both components enabled, demonstrating their synergistic effect. These results confirm the necessity of both temporal tracking and robust filtering in our framework.

\noindent\textbf{Sensitivity Analysis.}
We conduct a sensitivity analysis to assess how variations in the anomaly threshold \(\tau\) affect our framework's performance. Fig.~\ref{fig:experiments2} shows the impact of \(\tau\) on four metrics—Pixel-AUROC, Pixel-AP, TBDC, and RBDC—using the ShanghaiTech Campus dataset. As \(\tau\) ranges from 1.0 to 1.9, the framework performs stably, with optimal results at \(\tau = 1.6\). TBDC is most sensitive, peaking at 85\% at \(\tau = 1.6\), while RBDC stays around 62\%. Pixel-level metrics, Pixel-AUROC and Pixel-AP, also show steady trends, reflecting the framework's balance of fine-grained anomaly detection and object consistency. This analysis highlights the robustness and adaptability of our framework across threshold variations, ensuring reliable detection in diverse scenarios.

\noindent\textbf{Robustness Testing.}
We evaluate our framework's robustness through extensive testing across different SAM2 backbone architectures (Fig.~\ref{fig:experiments3}). Our analysis covers models from the lightweight Tiny variant to the full-scale Large model, with Base+ consistently yielding the best results. This stability suggests that the framework's effectiveness is not dependent on backbone capacity. While larger models show slight improvements in pixel-level metrics, performance remains strong across all variants. These results validate the framework’s reliability and robustness, demonstrating its suitability for deployment in a wide range of scenarios, irrespective of computational constraints.

\section{Conclusion}
\label{sec:conclusion}
This paper presents \model, a framework integrating coarse object detection with fine-grained anomaly localization in videos. Leveraging SAM2 and our dual-level evaluation, it enables precise segmentation and robust tracking. Experiments on UCSD Ped2 and ShanghaiTech datasets demonstrate state-of-the-art performance in pixel- and object-level metrics, advancing video anomaly detection.

{
\small
\bibliographystyle{ieeenat_fullname}

}

\end{document}